# DAFFNet: A Dual Attention Feature Fusion Network for Classification of White Blood Cells

Yuzhuo Chen[1#], Zetong Chen[1#], Yunuo An[1], Chenyang Lu[1], Xu Qiao[1*]

[1]School of Control Science and Engineering, Shandong University, Jinan City, Shandong Province, 250061, China

#These authors contributed to the work equally and should be regarded as co-first authors.

**Corresponding authors**: Xu Qiao

Xu Qiao

School of Control Science and Engineering, Shandong University

Address: Qianfoshan Campus, Shandong University, 17923 Jingshi Road, Jinan City, Shandong Province, 250061, China

Telephone number: +86-18663759381

E-mail: qiaoxu@sdu.edu.cn

# Abstract

The precise categorization of white blood cell (WBC) is crucial for diagnosing blood-related disorders. However, manual analysis in clinical settings is time-consuming, labor-intensive, and prone to errors. Numerous studies have employed machine learning and deep learning techniques to achieve objective WBC classification, yet these studies have not fully utilized the information of WBC images. Therefore, our motivation is to comprehensively utilize the morphological information and high-level semantic information of WBC images to achieve accurate classification of WBC. In this study, we propose a novel dual-branch network Dual Attention Feature Fusion Network (DAFFNet), which for the first time integrates the high-level semantic features with morphological features of WBC to achieve accurate classification. Specifically, we introduce a dual attention mechanism, which enables the model to utilize the channel features and spatially localized features of the image more comprehensively. Morphological Feature Extractor (MFE), comprising Morphological Attributes Predictor (MAP) and Morphological Attributes Encoder (MAE), is proposed to extract the morphological features of WBC. We also implement Deep-supervised Learning (DSL) and Semi-supervised Learning (SSL) training strategies for MAE to enhance its performance. Our proposed network framework achieves 98.77%, 91.30%, 98.36%, 99.71%, 98.45%, and 98.85% overall accuracy on the six public datasets PBC, LISC, Raabin-WBC, BCCD, LDWBC, and Labelled, respectively, demonstrating superior effectiveness compared to existing studies. The results indicate that the WBC classification combining high-level semantic features and low-level morphological features is of great significance, which lays the foundation for objective and accurate classification of WBC in microscopic blood cell images.

# 1. Introduction

White blood cells (WBCs) are hematopoietic cells originating from the bone marrow and present in the blood and lymphoid tissues. They serve as the cornerstone of the immune system, safeguarding the body against external pathogens and infectious agents[1][2][3]. The various types of white blood cells hold distinct clinical significance, with their quantities and ratios offering valuable insights into disease conditions and immune responses. Accurate classification of WBC is pivotal in the identification and management of blood-related disorders like leukemia, bacterial infections, and anemia[4][5]. Typically, medical professionals diagnose such conditions by visually examining and assessing the morphological, chromatic, and textural attributes of WBCs in blood or bone marrow specimens through microscopic analysis chromatic[6][7]. This diagnostic procedure is laborious, time-intensive, and heavily reliant on the proficiency of hematologists. Consequently, the contemporary focus is on the advancement of Computer-Aided Diagnostic (CAD) technologies to augment the capabilities of hematologists in clinical practice.

In recent years, studies on the development of CAD for WBC classification can be broadly categorized into two groups. The first category involves utilizing morphological features of WBC images, such as geometry, color, and texture features, as inputs for 1D machine learning models to facilitate classification. For instance, Agaian et al.[8] extracted shape features, color features, GLCM features and Haar wavelet features of leukocyte images and input them into SVM classifier to classify acute lymphoblastic leukemia and achieved 94% accuracy. Hegde et al.[9] extracted shape features and texture features of leukocyte nuclei and input them into hybrid SVM and NN classifier for leukocyte classification, achieving 99% accuracy. Dinčić et al.[10] quantified leukocyte structural and textural differences and selected eight features to input into an SVM classifier for classification based on the statistical ranges of the parameters of the different WBC types, achieving an average of 80% accuracy on average. Dasariraju et al.[11] extracted 16 features from leukocytes, and proposed two new nuclear color features, input features into RF classifier to classify immature leukocytes, which achieved 92.99% second classification accuracy.

Manually designed morphological features may not fully cover all cellular features, which cannot well represent the complex relationships and patterns in the images. In addition, morphological features are highly influenced by factors such as light, noise, staining method, imaging equipment, etc., which may lead to large differences in the extracted features when processing different types or sources of data. Consequently, the second category is directly based on the WBC images, which are input into an innovative neural network model to classify WBC using high-level deep learning features. For example, Jiang et al.[12] propose a deep neural network for WBC classification with DRFA-Net, which can accurately localize the WBC region and achieved five classification accuracies of 95.17%, 93.21%, and 95.87% on Raabin-WBC, LISC, and BCCD, respectively. Baydilli and Atila[13] used the capsule network to achieve 96.86% classification accuracy on the LISC dataset. Das and Meher[14] propose a deep CNN framework with a new weighting factor that achieved 99.39% and 97.18% accuracy on the ALLIDB1 and ALLIDB2 datasets, respectively. Wang et al.[15] propose the WBC-AMNet to automatically classify leukocyte subtypes based on a focalized attention mechanism, which achieved 95.66% accuracy on BCCD. Yan Ha et al.[16] come up with a novel semi-supervised leukocyte classification method named Fine-grained Interactive Attentional Learning (FIAL), which achieved 93.2% classification accuracy on the BCCD dataset.

Existing research has demonstrated the effectiveness of both high-level semantic features derived from deep learning models and morphological features of WBC in classification tasks. However, the current study does not integrate both high-level semantic features and morphological features of WBC. While extracting morphological features enables the model to effectively capture the relationship between morphological information and different categories, it may overlook image details crucial for classification. Moreover, relying solely on morphological features can lead to poor generalization due to their susceptibility to image quality variations. Conversely, another line of research involves directly feeding images into deep learning models for classification, allowing the model to learn complex high-level semantic information but potentially missing out on low-level morphological information, which are less interpretable. To enhance the representational capacity and accuracy of the model by leveraging both types of information present in WBC images, this study proposed a network that integrates high-level semantic features with morphological features in WBC images.

In this research, we introduce Dual Attention Feature Fusion Network (DAFFNet) as a novel approach to effectively integrate image morphological features and high-level semantic features. The network architecture is constructed based on ResNet and incorporates Dual Attention mechanisms through the integration of Efficient Pyramid Squeeze Attention (EPSA) and Spatial Channel (SA). Additionally, we introduce the Morphological Feature Extractor (MFE), comprising two key components: the Morphological Attributes Predictor (MAP) and the Morphological Attributes Encoder (MAE). To enhance the accuracy of morphological feature extraction from WBC images, a combination of Deep Supervision Learning (DSL) and Semi-supervised Learning (SSL) strategies was employed. Subsequently, the morphological features extracted by MFE are fused with the high-level semantic features obtained from the EPSA-SAResNet and inputted into the decoder for classification. This integration aims to fully leverage both low-level and high-level information present in the images. Through a series of ablation experiments conducted on each module of the proposed network framework, we demonstrate the reliability and robustness of our approach, establishing a solid foundation for the objective and accurate classification of WBC in microscopic blood cell images. The key contributions of this study are outlined as follows:

1) We propose a novel dual-branch network DAFFNet, which combines morphological features and high-level semantic features of WBC for the first time in the field of WBC classification. Comprehensively utilizing both low-level and high-level information in WBC images, we improve the model's characterization ability and classification performance.

2) Dual Attention mechanism is introduced by combining Efficient Pyramid Squeeze Attention (EPSA) and Spatial Channel (SA). The combined use of channel information and spatial information in the feature map enables the model to understand and utilize important features in a more comprehensive way.

3) The proposed MFE contains two modules, Morphological Attribute Predictor (MAP) and

Morphological Attribute Encoder (MAE). MAP can automatically and accurately extract morphological attributes, and MAE can encode the extracted attributes into higher-dimensional morphological features to enhance the model's ability to classify WBC.

4) DSL and SSL are used to train the MAP. DSL enables the MAP to fully utilize the sample category labels, while SSL enables the MFE to learn the category sample distribution in different datasets, achieving higher generalizability and accuracy.

5) Multi-center validation on six datasets is utilized to verify the robustness and effectiveness of the proposed model. The high-quality large-scale public dataset Labelled and the dataset containing morphological annotations WBCAtt are utilized for the first time.

## 2. Method

In this section, we introduce the network architecture of DAFFNet in detail, as shown in Figure 1. DAFFNet consists of two parallel feature extractors to extract high-level semantic features and morphological features and input the fused features into a decoder for classification. Among them, the high-level semantic feature extractor introduces EPSA and SA based on ResNet50. EPSA improves the network attention to the information of different channels in the input image, while SA enhances the network attention to the key regions in the image. the selection of ResNet50 as a backbone involved the comparison of ViT-B/16, ResNet50, and VGG19. The proposed MFE consists of two modules, MAP and MAE. MAP can automatically predict morphological attribute from WBC images, while MAE can encode the attribute into continuous morphological features. By fusing the high-level semantic features extracted by EPSA-SAResNet50 with the morphological features extracted by MFE, DAFFNet can comprehensively utilize the morphological features and high-level semantic features of WBC to achieve accurate classification of WBC.

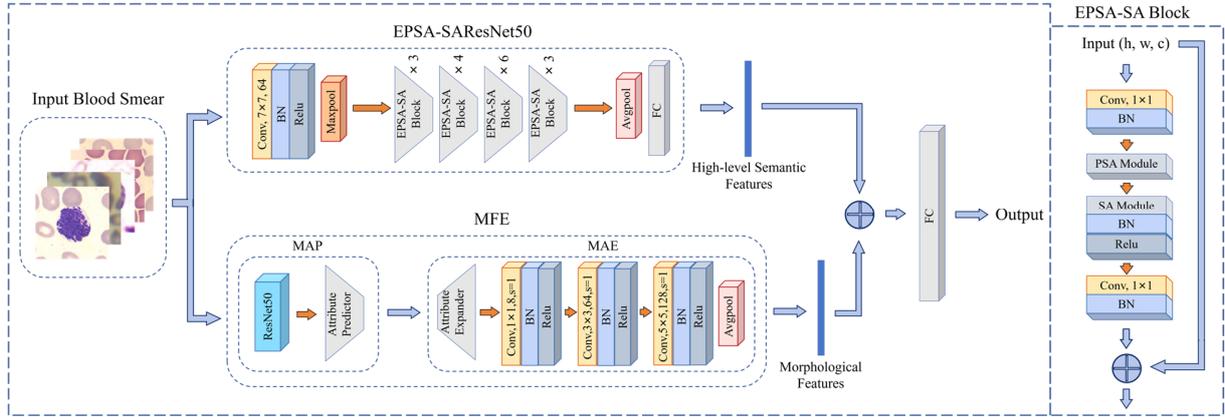

Figure 1. The structure of DAFFNet.

## 2.1. EPSA-SA

Currently, the attention mechanism is widely used in the field of image classification. For leukocyte microscopy images, the color and size of cells in different categories are different. In order to make the model better utilize the channel information and spatial information in the image, this study introduced dual attention mechanism by combining EPSA[17] and Spatial Attention (SA)[18]. EPSA can efficiently extract finer-grained multi-scale spatial information and form remote channel dependencies efficiently. It utilizes the multi-scale pyramid convolution structure to integrate the information of the input feature maps. At the same time, by squeezing the channel dimension of the input tensor, it can efficiently extract spatial information at different scales from each channel feature map. SA helps the model to better capture spatially localized features in an image by searching for important spatial regions in the image, and dynamically enhancing or weakening the weight of the features in a specific region. EPSA-SA can comprehensively utilize the channel information and spatial information in the feature map, which enables the model to understand and utilize the features more comprehensively. In this study, EPSA-SAResNet can capture high-level and fine-grained details to classify WBC in blood smear images, thus improving the accuracy of classification.

## 2.2. Morphological Feature Extractor (MFE)

Existing research has demonstrated the efficacy of high-level semantic features and morphological characteristics of WBC in classification tasks. To improve the model classification performance by utilizing multi-level features, MFE was introduced in this study to extract morphological features from WBC images. MFE consists of two modules, MAP and MAE. The MAP module predicts the morphological attributes of WBC, while the MAE module transforms these attributes into higher dimensional morphological features. In this subsection, we describe the architecture and training strategy of each module of MFE in detail.

*2.2.1. Morphological Attribute Predictor (MAP)*

MAP can be regarded as a multi-label, multi-class predictor. MAP utilize ResNet50 as the front-end and the tail-end maps the features of [1000, 1] to 11 attributes through 11 parallel Fully Connected (FC) layers. Each attribute has multiple categories with dimension [number of classes, 1]. The prediction of categories for each attribute can be obtained by performing argmax independently for each attribute.

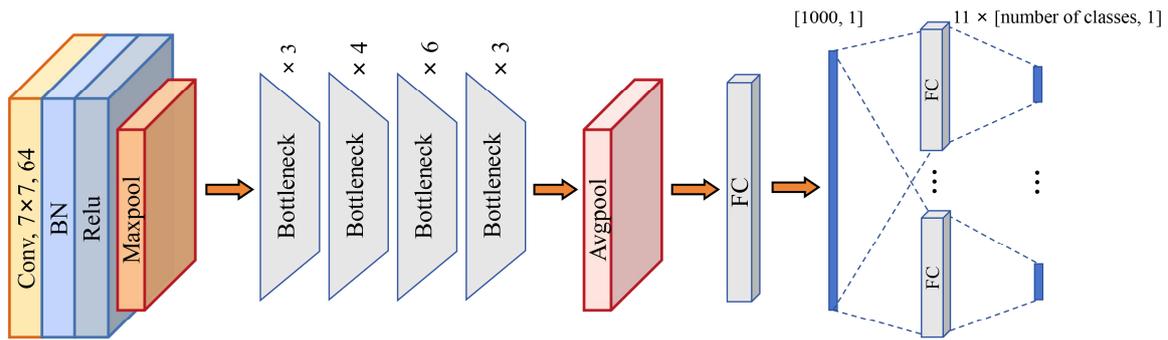

Figure 2. The structure of MAP.

*2.2.2. Morphological Attribute Encoder (MAE)*

The utilization of the 11 morphological attributes as classification features presents several challenges. Firstly, due to their discrete integer nature, morphological attributes are limited in their ability to effectively characterize features and cannot represent continuous transition information. For instance, the morphological attributes of cell size are only large and small, but cells come in a variety of sizes and should not be measured simply by two discrete values. When using morphological attributes as features, the model not only fails to capture the complex patterns of the dataset, but also causes a large accumulation of systematic errors in the feature extraction module, which leads to a degradation of classification performance. Secondly, the number of

11 morphological features is too small, making the feature imbalance problem serious when fused with the 1000 high-level semantic features. Therefore, MAE was proposed in this study. With MAE, the 11 parallel attribute probability values predicted by MAP can be encoded into a continuous-valued morphological feature encoding vector of length 128.

As shown in Figure 3, the MAE first expands each attribute vector from the dimensional prediction of [number of classes, 1] to [10 × number of classes, 1] by FC layers. After the input sigmoid is activated, it is then squeezed to [11, 1]. The eleven vectors are concatenated along dimension 0 to generate a feature map of dimension 11 × 11. The feature map is sequentially passed through three convolutional layers (kernel size=1 × 1, 3 × 3, 5 × 5, stride=1,1,1, out channel=8, 64, 128). Batch normalization and ReLU activation operations are applied after each convolution. Finally, the generated features are subjected to average pooling operation to obtain a morphological feature coding vector of length 128.

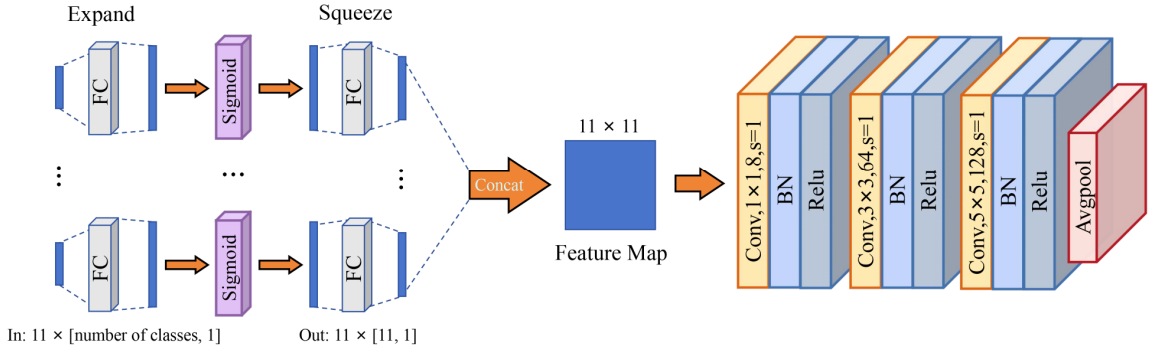

Figure 3. The structure of MAE.

### 2.2.3. Deep and Semi-supervised Learning

Different types of WBC exhibit distinct distributions of morphological features. To optimize the utilization of category labels and enhance the efficacy of MAP, the DSL strategy was implemented. We incorporated a linear layer at the end of MAP to classify the extracted features into cell categories. By integrating the classification loss with the attribute prediction loss, the model can learn the distribution of morphological features of different categories. The specific loss function utilized in this study is represented by equation (1):

$$\text{Loss} = -\sum_{i=1}^{N} \left[ \lambda_{AP} \frac{\sum_{m=1}^{A}\sum_{t=1}^{P} y_{imt} \log_2 \text{map}(x_i)_{mt}}{A} + \lambda_{Cls} \sum_{c=1}^{K} y_{ic} \log_2 \text{mfec}(x_i)_c \right] \quad (1)$$

Where $\lambda_{AP}$ denotes the weight of the attribute predictor and $\lambda_{Cls}$ denotes the weight of the auxiliary classifier. Based on experience, we select $\lambda_{AP}$ as 0.8 and $\lambda_{Cls}$ as 0.2. N denotes the number of samples. A denotes the number of WBC morphological attributes. P denotes the number of categories of a particular morphological attribute. K denotes the number of WBC categories. $y_{imt}$ denotes the one-hot encoding of WBC attribute m category t. $y_{ic}$ denotes the one-hot encoding of the target values of WBC category c coding. $map(x_{im})_t$ denotes the probability that the attribute m of the model-predicted sample $x_i$ belongs to category t. $mfec(x_i)_c$ denotes the probability that the MFE-predicted sample $x_i$ with a categorization layer added to the tail belongs to category c. $mfec(x_i)_c$ denotes the probability that the model-predicted sample $x_i$ belongs to category c.

Among the six datasets collected in this study, only the PBC dataset exists the morphological attribute annotation. To make full use of the data in multiple datasets and to improve the generalization and characterization of MFE, this study also utilized the SSL[19] strategy. First, the MAP was pretrained using the annotations in the WBCAtt dataset for the WBC morphological attributes of the PBC dataset. Then, the trained MAP was utilized to predict the morphological attributes of the WBC images from the training sets of the other five datasets to obtain pseudo-label. Finally, the true-label from PBC was mixed with the pseudo-label from the training sets of the other five datasets, and the MAP was trained for the second time using the images from all datasets. By integrating the DSL with the SSL, the study aimed to leverage all available sample data category labels to enable the MAP to learn diverse category distributions and sample distributions, thereby enhancing generalization and accuracy.

Figure 4 shows how we trained MAP using DSL with SSL.

Step 1. Training MAP by Deep-supervised Learning(DSL) based on PBC and WBCAtt datasets.

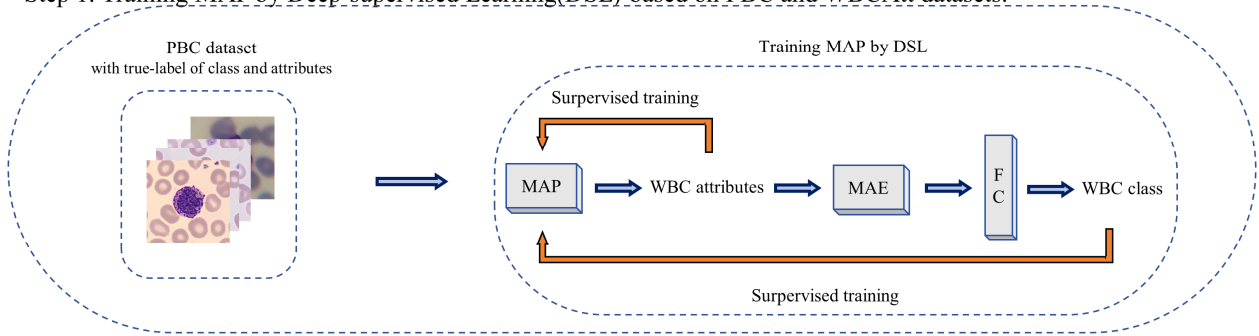

Step 2. Predicting the morphological attributes of WBC in other five datasets by using MAP trained by DSL.

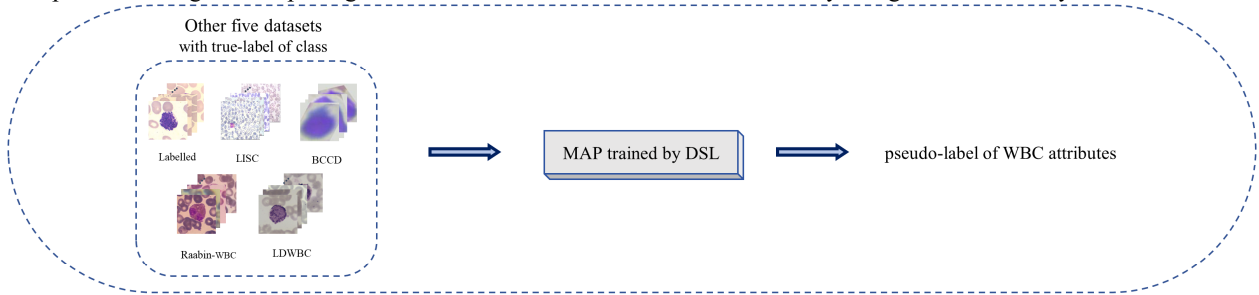

Step 3. Retraining MAP by DSL based on all datasets with their true-label of class and mixed-label of attributes.

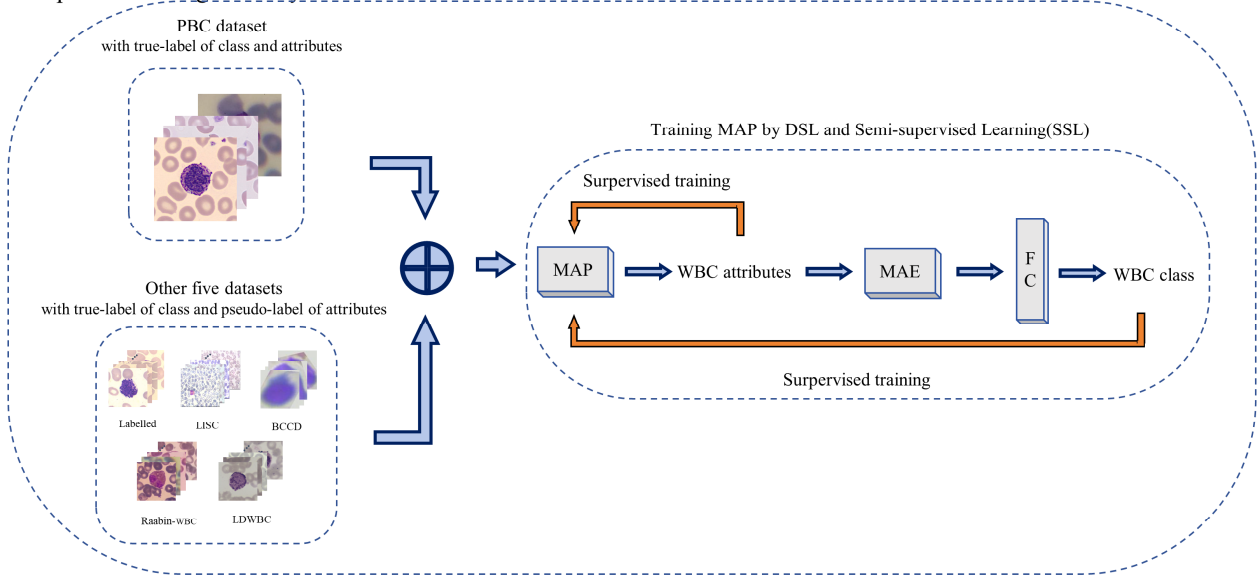

Figure 4. Steps for training MAP modules using DSL with SSL

## 3. Experiment

### 3.1. Data and setting

    To verify the effectiveness and generalizability of the model and enhance the credibility of the experimental results, we validated our results in six representative WBC classification datasets, namely, PBC, LISC, Raabin-WBC, BCCD, LDWBC, and Labelled. In addition, for training MAP, we also collected datasets

with WBC morphological attributes labeled annotations. For each dataset, we randomly divided the training set, validation set, and test set in a ratio of 6:2:2. A description of the datasets used is shown in Table 1, and a picture of the WBC categories in the five categories of routine blood is shown in Figure 5. The number of cells in each category in the used dataset is shown in Supplementary Table 1. Additionally, a detailed explanation of the information not included in the table is provided below:

The PBC dataset was collected from subjects who did not have infections, hematological or neoplastic diseases, and were not receiving any medications at the time of blood collection. The LISC dataset is an older leukocyte dataset consisting of blood images from peripheral blood samples of healthy individuals. The Raabin-WBC dataset was collected by Iranian medical laboratories from normal peripheral blood slides from 72 male and female subjects aged 12 to 70 years. In addition, basophil images were collected from chronic granulocytic leukemia (CML)-positive samples. The Raabin-WBC dataset was collected from 72 normal peripheral blood samples of male and female subjects between the ages of 12 and 70 years from Iranian medical laboratories, in addition to basophil images from chronic granulocytic leukemia (CML)-positive samples. The LDWBC dataset was collected from 150 microscopic slide samples from 150 normal individuals. The images in the Labelled dataset also capture blood film artifacts that can occur under various lighting conditions and during manual preparation and staining of blood smears. The LDWBC dataset uses a high-quality acquisition device to collect blood film images of individuals. The dataset uses high-quality acquisition equipment and has a resolution of approximately 42 pixels per 1μm per image, which is the highest quality of blood cell images obtained from manually stained blood films under a microscope. The WBCAtt dataset[20] identified 11 morphological attributes associated with cells and their components (nuclei, cytoplasm, and granules) through collaboration with pathologists, a thorough literature review, and manual examination of microscopic images. The categories included in each attribute shows in Supplementary Table 2. These attributes were then used to annotate more than 10,000 WBC images in the PBC dataset, yielding 113k tags (11 attributes × 10.3k images). This level of detail and scale of annotation is unprecedented and provides unique value for artificial intelligence in pathology. In this study, MFE is trained using WBCAtt in conjunction with PBC dataset to enhance the model performance.

| DATASET | Total sample number | Image size(px) | Image format | Stain method | Imaging device | Area |
|---|---|---|---|---|---|---|
| PBC[21] | 17092 | (360, 363) | JPG | May Grünwald Giemsa | CellaVision DM96 | Barcelona, Spain |
| LISC[22] | 242 | (720, 576) | BMP | Gismo Right | SSC-DC50AP(Sony) & Axioskope 40 microscope | Tehran, Iran |
| Raabin-WBC[23] | 14514 | (575, 575) | JPG | Giemsa | Olympus CX18 & Zeiss at a magnification of 100× | Tehran, Iran |
| BCCD[24] | 3500 | (84, 84) | JPG | Gismo Right | General optical microscope with 100× objective & Analogue CCD colour camera | America |
| LDWBC[25] | 22645 | (1280, 1280) | JPG | Wright Giemsa | OLYMPUS BX41 & OLYMPUS Plan N 100x/1.25 | Wuhan, China |
| Labelled[26] | 16027 | (1200, 1200) | BMP | May Grünwald & Giemsa Romanowski | Olympus BX51 microscope & Basler acA5472-17uc | Ostrava, Czech Republic |

Table 1.Basic information about each dataset.

| Class | PBC | LISC | Raabin-WBC | BCCD | LDWBC | Labelled |
|---|---|---|---|---|---|---|
| Basophil | 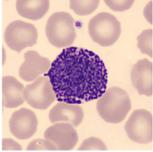 | 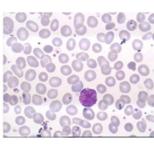 | 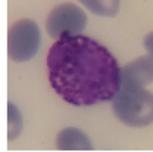 | 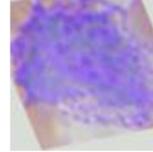 | 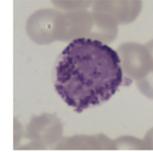 | 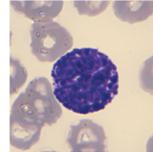 |
| Eosinophil | 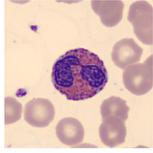 | 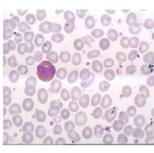 | 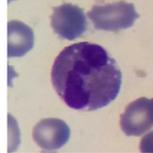 | 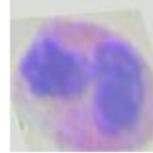 | 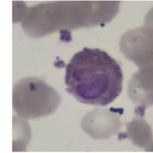 | 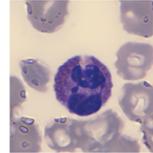 |
| Lymphocyte | 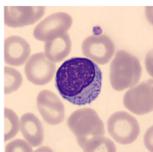 | 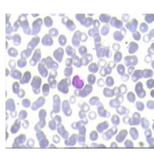 | 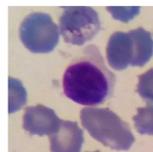 | 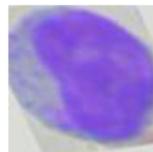 | 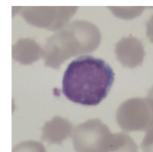 | 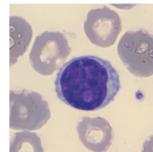 |
| Monocyte | 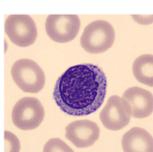 | 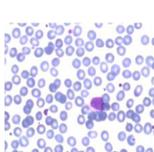 | 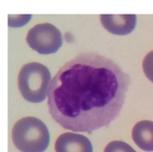 | 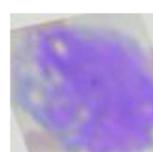 | 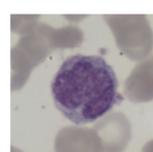 | 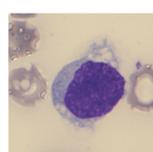 |
| Neutrophil | 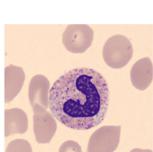 | 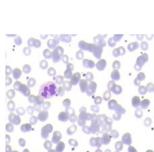 | 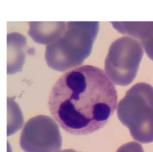 | 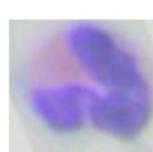 | 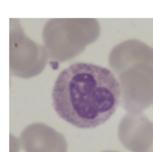 | 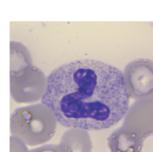 |

Figure 5. Pictures of WBCs by category in the five categories of routine blood in the dataset used.

## 3.2. Experiment details

In this study we implemented a classification model using the Pytorch deep learning framework in Python version 3.8, running on a TITAN RTX with 24GB of memory. The CUDA version utilized was 11.8. The model employed ResNet50 as backbone with a transfer learning training strategy, initializing the model weights pre-trained on ImageNet. A freeze weight training strategy was applied for fine-tuning during the training process. The training was conducted over 100 epochs, with early stopping implemented using a patience value of 10 to prevent overfitting. The optimizer chosen for the network was Adam, with a batch size of 32 and a learning rate of 0.0001. Furthermore, the input training set images were randomly cropped to

224×224, while the validation set images were center cropped to the same dimensions. The model exhibiting the best performance on the validation set was selected for evaluation on the test set.

### 3.3. Model evaluation parameters

To validate the effectiveness of the model, we introduced various evaluation metrics to assess the model performance comprehensively and objectively. These include Precision (Prec), Recall (Rec), F1-score (F1), Accuracy (Acc), Area Under the Curve (AUC). For Prec, Rec, and F1, we take the weighted average value. In addition, we introduce Subset Accuracy (SAcc), Hamming Loss (HL), Jaccard Similarity (JS) to evaluate the prediction performance of MAP , which can be considered as a multi-label multi-categorization predictor. The formulas for some of these evaluation metrics are shown in Eqs. (2)-(5):

$$weighted\ Avg = \sum_{i=0}^{classes\ number} class\ i\ metrics \times \frac{class\ i\ number}{all\ class\ number} \tag{2}$$

$$SAcc = \frac{1}{m}\sum_{i=1}^{m} I(y^{(i)} == \hat{y}^{(i)}) \tag{3}$$

$$HL = \frac{1}{mq}\sum_{i=1}^{m}\sum_{j=1}^{q} I(y^{(i)} \neq \hat{y}^{(i)}) \tag{4}$$

$$JS = \frac{|(y^{(i)} \cap \hat{y}^{(i)})|}{|(y^{(i)} \cup \hat{y}^{(i)})|} \tag{5}$$

Among them, $y^{(i)}$ is true labels, $\hat{y}^{(i)}$ is prediction results. SAcc is the ratio of samples in which the predicted value is exactly the same as the true value in all samples. HL is the ratio of wrongly predicted labels in all samples. JS is the Ratio of the intersection between $y^{(i)}$ and $\hat{y}^{(i)}$ to the concatenation between $y^{(i)}$ and $\hat{y}^{(i)}$.

### 3.4. Experiment Results

#### 3.4.1. Classification performance

The multicenter validation results of the proposed model DAFFNet on six datasets, PBC, LISC, Raabin-WBC, BCCD, LDWBC, and Labelled, are shown in Table 2. Among them, the model has the highest overall metrics on the BCCD dataset, with Prec, Rec, F1, Acc, and AUC reaching 99.72%, 99.71%, 99.71%, 99.71%, 99.82%, and the overall metrics reaching state-or-the-art. This result highlights the superiority of DAFFNet, and proves the effectiveness of the proposed framework. On the three datasets of Raabin-WBC, LDWBC, and Labelled, the Prec, Rec, F1, Acc, and AUC of DAFFNet also exceed 98%, underscoring its robustness and effectiveness. Even on the LISC dataset, which comprises only 242 samples, DAFFNet exhibited commendable overall metrics, with Precision, Recall, F1 score, Accuracy, and AUC reaching 92.49%, 91.30%, 90.92%, 91.30%, and 94.52%, respectively. These findings suggest that DAFFNet is less dependent on the amount of data, thereby laying a foundation for its practical implementation.

| Method | Dataset | Prec | Rec | F1 | Acc | AUC |
| --- | --- | --- | --- | --- | --- | --- |
| DAFFNet | PBC | 98.77% | 98.77% | 98.77% | 98.77% | 99.28% |
| | LISC | 92.49% | 91.30% | 90.92% | 91.30% | 94.52% |
| | Raabin-WBC | 98.41% | 98.36% | 98.38% | 98.36% | 98.91% |
| | BCCD | 99.72% | 99.71% | 99.71% | 99.71% | 99.82% |
| | LDWBC | 98.44% | 98.45% | 98.44% | 98.45% | 98.82% |
| | Labelled | 98.81% | 98.85% | 98.81% | 98.85% | 99.32% |

Table 2. results of the proposed DAFFNet on six datasets: PBC, LISC, Raabin-WBC, BCCD, LDWBC, Labelled.

The confusion matrix of DAFFNet's prediction results for the test set in each dataset is shown in Figure 6. It can be seen that the LDWBC, Raabin-WBC, and Labelled datasets exhibit notable data imbalance issues. For example, in the prediction set of Raabin-WBC, there are only 80, 106 and 161 samples for Basophil, Eosinophil and Monocyte respectively, while there are 2066 and 2083 samples for Lymphocyte and Neutrophil. However, for Basophil, Eosinophil, and Monocyte, which have very few samples, there are still good prediction results. This highlights the robustness and effectiveness of DAFFNet in handling imbalanced data, thereby establishing its potential utility in clinical medical settings.

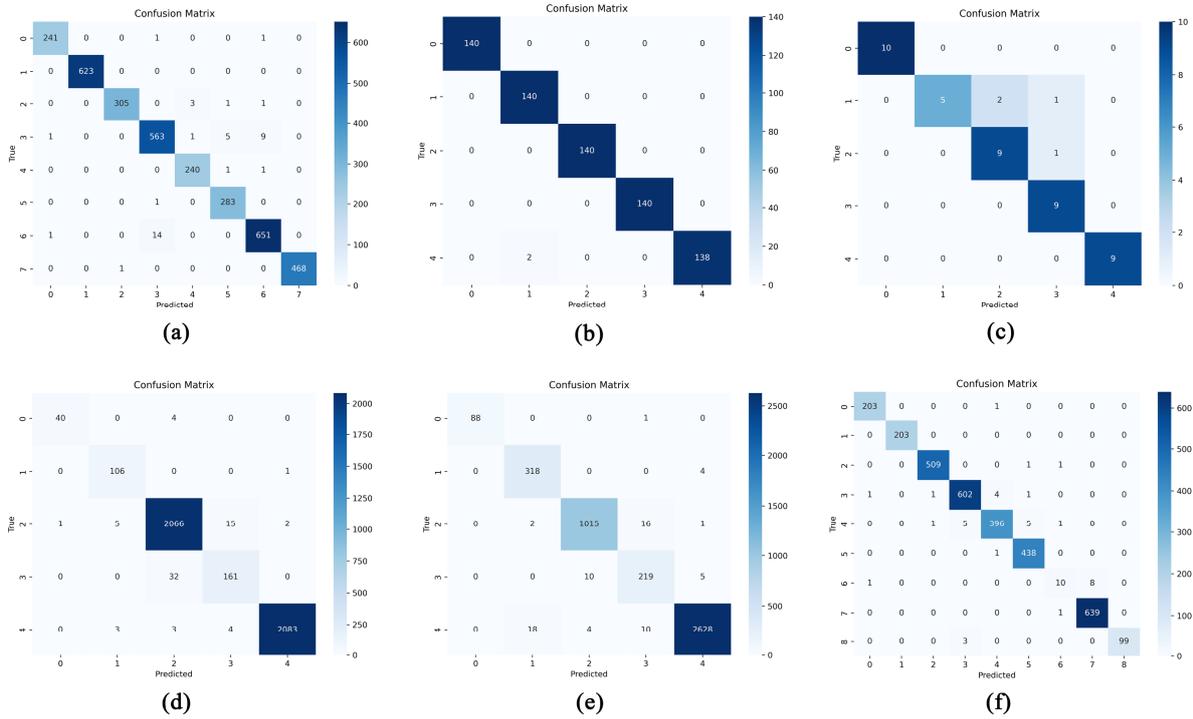

Figure 6.Comparison of the proposed DAFFNet with existing methods, in terms of overall accuracy, precision, recall, F1-Score, AUC on PBC, LISC, Raabin-WBC, BCCD, LDWBC, and Labelled benchmark datasets.

### 3.4.2. Results of comparison with recent advances

In order to further demonstrate the superiority of our proposed method, a comparative analysis was conducted between the outcomes of our method and t other studies in the field of WBC classification in recent years.These studies have focused on either high-level semantic features or morphological features of WBC and have not comprehensively utilize the information from WBC images. All comparison experiments were conducted with the same dataset and multiple evaluation metrics were used for a comprehensive comparison. The results are shown in Table 3, which describes the performance metrics of the different studies in the PBC, LISC, Raabin-WBC, BCCD, LDWBC, Labelled.

The results show that while previous techniques have achieved high classification accuracy, our proposed method demonstrates superior efficacy. Notably, our method outperforms most of the existing methods for leukocyte classification on all six datasets, confirming the effectiveness of the findings and the generalization ability of the model. In addition, our model achieved 99.72% precision, 99.71% recall, 99.71% F1-score, 99.71% accuracy, and 99.82% AUC on the BCCD test set, with an overall metric of state-of-the-art.

| Dataset | Reference | Year | Prec | Rec | F1 | Acc | AUC |
|---|---|---|---|---|---|---|---|
| PBC | Dhar, P.[27] | 2022 | 97.81% | 98.10% | 97.95% | - | - |
| | Tseng, T. R.[28] | 2023 | 85.10% | - | - | 90.10% | - |
| | Jimut Bahan Pal[29] | 2024 | 95.64% | 95.77% | 95.80% | 96.32% | - |
| | Proposed DAFFNet | - | **98.77%** | **98.77%** | **98.77%** | **98.77%** | **99.28%** |
| LISC | Tavakoli E[30] | 2021 | - | - | - | 88.21% | - |
| | Safuan, S. N. M.[31] | 2022 | - | - | - | 84.52% | - |
| | Luu, V. Q[32] | 2024 | - | - | - | 76.49% | - |
| | Proposed DAFFNet | - | **92.49%** | **91.30%** | **90.92%** | **91.30%** | **94.52%** |
| Raabin-WBC | Tavakoli, S.[33] | 2021 | - | - | - | 94.65% | - |
| | Rivas-Posada, E.[34] | 2021 | 95.65% | - | 96.77% | 97.69% | - |
| | Jiang[12] | 2022 | 90.43% | - | 91.89% | 95.17% | - |
| | Proposed DAFFNet | - | **98.41%** | **98.36%** | **98.38%** | **98.36%** | **98.91%** |
| BCCD | Ha, Y.[16] | 2022 | 93% | 93% | 93.50% | 93.22% | - |
| | Rao, B. S. S.[35] | 2023 | 99.22% | 99.10% | 99.18% | 99.19% | - |
| | Balasubramanian, K.[36] | 2023 | - | - | - | 98.69% | - |
| | Proposed DAFFNet | - | **99.72%** | **99.71%** | **99.71%** | **99.71%** | **99.82%** |
| LDWBC | Chen, H.[37] | 2022 | 93.51% | 95.06% | 94.17% | 98.06% | - |
| | Proposed DAFFNet | - | **98.44%** | **98.45%** | **98.44%** | **98.45%** | **98.82%** |
| Labelled | Bodzas, A.[26] | 2023 | - | - | - | 94.49% | - |
| | Proposed DAFFNet | - | **98.81%** | **98.85%** | **98.81%** | **98.85%** | **99.32%** |

Table 3.Comparison of the proposed DAFFNet with existing methods, in terms of overall accuracy, precision, recall, F1-Score, AUC on PBC, LISC, Raabin-WBC, BCCD, LDWBC, and Labelled benchmark datasets.

### 3.4.3. Ablation experiments results

Ablation experiment is a scientifically valid method of experimental design whose main purpose is to test hypotheses by systematically adding, subtracting, or changing certain factors in a study to assess the effects of these factors on the results. In order to validate and reveal the effects of different modules in DAFFNet on the original baseline ResNet50, this study evaluated the EPSA Attention Module, the SA Attention Module, the MAP Module, and the MAE Module through ablation experiments.

In ablation experiments, multifactorial experiments are combined improvements to demonstrate the validity of the overall performance. In each of the six datasets, PBC, LISC, Raabin-WBC, BCCD, LDWBC, and Labelled, additional improvements were made to the previous set of experiments. The experimental results are shown in Table 4. As can be seen from the table, compared to baseline, the classification performance of model after the introduction of EPSA increases from 98.62%, 97.14%, 97.86%, 97.81%, 96.84% to 98.65%, 97.42%, 99.14%, 97.99%, 98.18% on PBC, Raabin-WBC, BCCD, LDWBC, Labelled. On this basis, the results of introducing the dual attention combining EPSA and SA on the five datasets of PBC, LISC, Raabin-WBC, BCCD, and Labelled are all improved, proving the effectiveness of the proposed dual attention. In particular, the result of the model containing the four combined improvements mentioned above, the final DAFFNet, achieves the best results on each dataset. The Acc in PBC, LISC, Raabin-WBC, BCCD, LDWBC, and Labelled increased from 98.62%, 89.36%, 97.14%, 97.86%, 97.81%, and 96.84% in baseline to 98.77%, 91.30%, 98.36%, and 99.71%, 98.45%, and 98.85%. The results show that the proposed modules and improvements can effectively improve the performance of the model, which fully proves the effectiveness of the method.

| Dataset | Method | Prec | Rec | F1 | Acc | AUC |
|---|---|---|---|---|---|---|
| PBC | ResNet50 | 98.63% | 98.62% | 98.62% | 98.62% | 99.19% |
| | ResNet50+EPSA | 98.66% | 98.65% | 98.65% | 98.65% | 99.22% |
| | ResNet50+EPSA+SA | 98.68% | 98.68% | 98.68% | 98.68% | 99.24% |
| | <u>ResNet50+EPSA+SA+MAP</u> | <u>98.74%</u> | <u>98.74%</u> | <u>98.74%</u> | <u>98.74%</u> | <u>99.26%</u> |
| | **ResNet50+EPSA+SA+MAP+MAE** | **98.77%** | **98.77%** | **98.77%** | **98.77%** | **99.28%** |
| LISC | ResNet50 | 88.90% | 87.23% | 86.90% | 87.23% | 92.19% |
| | ResNet50+EPSA | 85.88% | 85.11% | 84.99% | 85.11% | 91.01% |
| | ResNet50+EPSA+SA | 88.90% | 87.23% | 86.90% | 87.23% | 92.19% |
| | <u>ResNet50+EPSA+SA+MAP</u> | <u>88.29%</u> | <u>89.36%</u> | <u>88.37%</u> | <u>89.36%</u> | <u>93.42%</u> |
| | **ResNet50+EPSA+SA+MAP+MAE** | **92.49%** | **91.30%** | **90.92%** | **91.30%** | **94.52%** |
| Raabin-WBC | ResNet50 | 97.31% | 97.14% | 97.16% | 97.14% | 98.04% |
| | ResNet50+EPSA | 97.49% | 97.42% | 97.38% | 97.42% | 97.99% |
| | ResNet50+EPSA+SA | 97.79% | 97.76% | 97.77% | 97.76% | 98.10% |
| | <u>ResNet50+EPSA+SA+MAP</u> | <u>98.09%</u> | <u>98.09%</u> | <u>98.05%</u> | <u>98.09%</u> | <u>98.58%</u> |
| | **ResNet50+EPSA+SA+MAP+MAE** | **98.41%** | **98.36%** | **98.38%** | **98.36%** | **98.91%** |
| BCCD | ResNet50 | 97.87% | 97.86% | 97.86% | 97.86% | 98.66% |
| | ResNet50+EPSA | 99.15% | 99.14% | 99.14% | 99.14% | 99.46% |
| | ResNet50+EPSA+SA | 99.44% | 99.43% | 99.43% | 99.43% | 99.64% |
| | <u>ResNet50+EPSA+SA+MAP</u> | <u>99.58%</u> | <u>99.57%</u> | <u>99.57%</u> | <u>99.57%</u> | <u>99.73%</u> |
| | **ResNet50+EPSA+SA+MAP+MAE** | **99.72%** | **99.71%** | **99.71%** | **99.71%** | **99.82%** |
| LDWBC | ResNet50 | 97.92% | 97.81% | 97.85% | 97.81% | 98.48% |
| | ResNet50+EPSA | 97.99% | 97.99% | 97.97% | 97.99% | 98.40% |
| | ResNet50+EPSA+SA | 97.81% | 97.88% | 97.83% | 97.88% | 98.32% |
| | <u>ResNet50+EPSA+SA+MAP</u> | <u>98.38%</u> | <u>98.28%</u> | <u>98.31%</u> | <u>98.28%</u> | <u>98.83%</u> |
| | **ResNet50+EPSA+SA+MAP+MAE** | **98.44%** | **98.45%** | **98.44%** | **98.45%** | **98.82%** |
| Labelled | ResNet50 | 96.82% | 96.84% | 96.79% | 96.84% | 98.15% |
| | ResNet50+EPSA | 98.21% | 98.18% | 98.00% | 98.18% | 98.91% |
| | ResNet50+EPSA+SA | 98.74% | 98.72% | 98.64% | 98.72% | 99.22% |
| | <u>ResNet50+EPSA+SA+MAP</u> | <u>98.77%</u> | <u>98.82%</u> | <u>98.76%</u> | <u>98.82%</u> | <u>99.31%</u> |
| | **ResNet50+EPSA+SA+MAP+MAE** | **98.81%** | **98.85%** | **98.81%** | **98.85%** | **99.32%** |

Table 4. The results of the ablation experiments for each module in DAFFNet. The added modules include EPSA, SA, MAP, MAE. The results show the improvement of each module on the performance of the classification network. (In the table, the highest performance model and its metrics are bolded, and the next highest performance model and its classification metrics are underlined).

### 3.4.4. Morphological attribute prediction performance

The prediction accuracy of morphological attributes is crucial for classification when using morphological features. The prediction outcomes of MAP are presented in Table 5, showcasing the performance metrics for 11 morphological attributes. Notably, the cumulative values for overall Prec, Rec, F1, Spec, Acc, SAcc, HL, and JS reached 85.56%, 84.20%, 95.10%, 84.24%, 88.65%, 89.65%, 35.37%, 10.41%, and 80.05%, respectively. This result highlights the effectiveness of MAP and lays the foundation for subsequent classification by combining morphological features. An overall high level of accuracy was attained across all eleven morphological attributes, underscoring the reliability of MAP.

| Method | Attribute | Prec | Rec | F1 | Spec | Acc | AUC | SAcc | HL | JS |
|---|---|---|---|---|---|---|---|---|---|---|
| MAP | Cell size | 81.19% | 77.45% | 82.37% | 79.28% | 79.93% | 79.91% | 35.37% | 10.41% | 80.05% |
| | Cell shape | 97.11% | 83.71% | 99.32% | 89.91% | 95.97% | 91.51% | | | |
| | Nucleus shape | 81.75% | 81.83% | 95.95% | 81.68% | 81.83% | 88.89% | | | |
| | Nuclear Cytoplasmic ratio | 91.60% | 97.96% | 98.79% | 94.67% | 98.69% | 98.37% | | | |
| | Chromatin density | 73.43% | 83.52% | 97.07% | 78.15% | 95.87% | 90.29% | | | |
| | Cytoplasm vacuole | 93.48% | 79.63% | 99.53% | 86.00% | 97.96% | 89.58% | | | |
| | Cytoplasm texture | 89.17% | 93.48% | 97.14% | 91.27% | 96.40% | 95.31% | | | |
| | Cytoplasm colour | 92.60% | 88.24% | 97.77% | 85.82% | 88.24% | 93.00% | | | |
| | Granule type | 68.55% | 67.83% | 88.97% | 67.59% | 67.83% | 78.40% | | | |
| | Granule colour | 73.06% | 72.59% | 89.41% | 72.63% | 72.59% | 81.00% | | | |
| | Granularity | 99.24% | 100.00% | 99.74% | 99.62% | 99.81% | 99.87% | | | |
| | **Overall** | **85.56%** | **84.20%** | **95.10%** | **84.24%** | **88.65%** | **89.65%** | | | |

Table 5. Test results of MAP trained by DSL and SSL strategies for each morphological attribute.

Figure 7 shows the confusion matrix of test results for each morphological attribute using MAP trained by DSL and SSL strategies. The analysis reveals a prevalent issue of sample imbalance across most attributes. For instance, the Nuclear Cytoplasmic ratio attribute exhibits a significant disparity with 1813 low samples and only 245 high samples. Notably, among the 245 high samples, only 5 are misclassified, indicating the model's capability in predicting attributes with imbalanced samples. The majority of morphological attributes demonstrate superior prediction accuracy, notably the Granularity attribute with 2058 samples, where only four samples were inaccurately predicted, underscoring the efficacy of MAP.

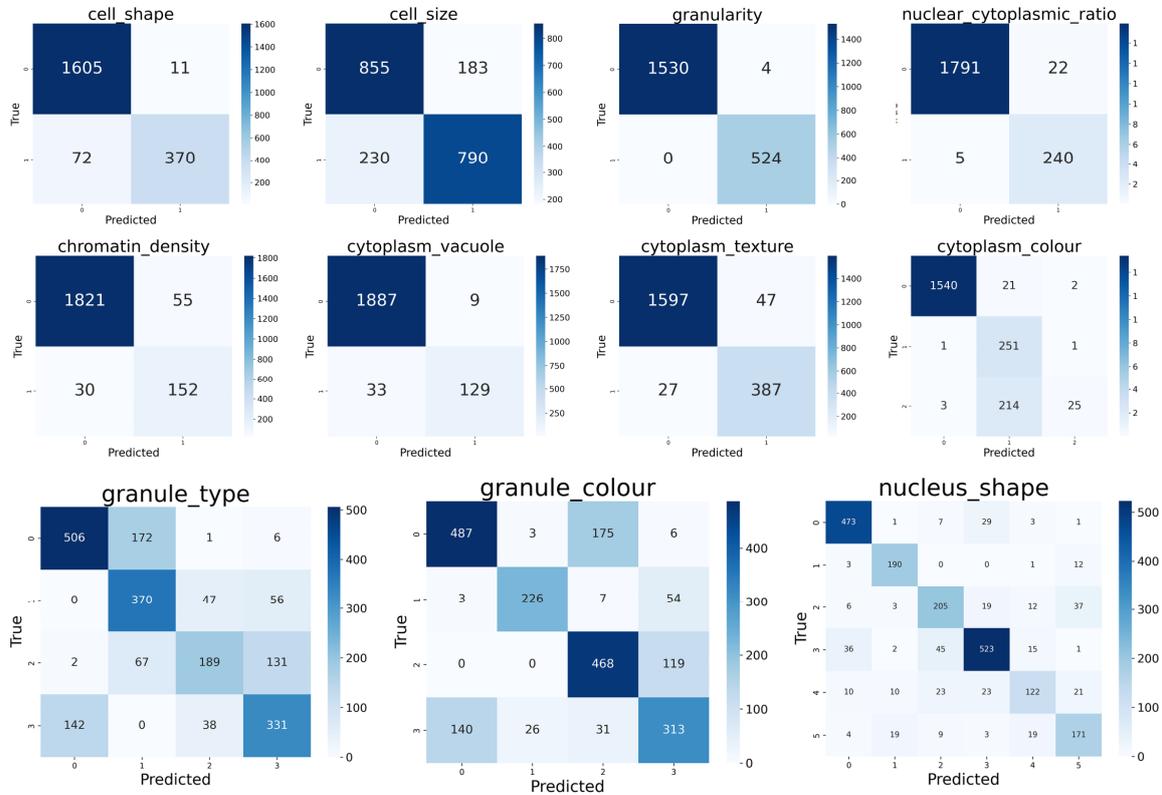

Figure 7. Confusion matrix of test results of MAP trained by DSL and SSL strategies for 11 morphological attributes.

To verify the effectiveness of DSL and SSL for MAP, an evaluation of its performance under various training strategies was conducted and the results are presented in Table 6. Upon implementing the DSL approach, which involves incorporating auxiliary classifiers into the MAP model, enhancements were observed across multiple evaluation metrics compared to the original MAP module. Specifically, Prec, Rec, F1, Spec, Acc, AUC, SAcc, and JS improve from 79.80%, 78.09%, 92.41%, 78.26%, 84.00%, 85.25%, 26.43%, and 70.41% to 83.11%, 81.94%, 94.29%, 81.92%, 87.05%, 88.12%, 29.54%, and 76.09%, respectively, and HL decreases from 14.67% to 11.87%, which verified the effectiveness of DSL. Furthermore, by leveraging SSL in conjunction with DSL to fully exploit all available sample data, a subsequent enhancement in the performance of MAP can be observed. Prec, Rec, F1, Spec, Acc, AUC, SAcc, and JS of MAP improve from 83.11%, 81.94%, 94.29%, 81.92%, 87.05%, 88.12%, 29.54% and 76.09% to 85.56%, 84.20%, 95.10%,

84.24%, 88.65%, 89.65%, 35.37% and 80.05%, respectively, and HL decreases from 11.87% to 10.41%. This result confirms that the combined utilization of DSL and SSL techniques empowers the MAP model to effectively leverage sample data and associated category labels, thereby enhancing its predictive capabilities. Additionally, the test results for each morphological attribute under different training strategies shows in Supplementary Table 3.

| Training Strategy | Prec | Rec | F1 | Spec | Acc | AUC | SAcc↑ | HL↓ | JS↑ |
|---|---|---|---|---|---|---|---|---|---|
| Self-supervised Learning | 79.80% | 78.09% | 92.41% | 78.26% | 84.00% | 85.25% | 26.43% | 14.67% | 70.41% |
| Deep-supervised Learning | 83.11% | 81.94% | 94.29% | 81.92% | 87.05% | 88.12% | 29.54% | 11.87% | 76.09% |
| Deep and Simi-supervised Learning | 85.56% | 84.20% | 95.10% | 84.24% | 88.65% | 89.65% | 35.37% | 10.41% | 80.05% |

Table 6. Test results of MAP when using different training strategies.

## 4. Discussion

Recent studies in the field of WBC classification can be broadly divided into two main categories. The first approach involves classifying WBC based on their morphological features, utilizing a one-dimensional machine learning model that considers surface features like geometry, color, and texture. The second approach involves directly input images of WBC into advanced neural network models to classify them based on high-level semantic features. While existing methods have demonstrated high accuracy, they cannot fully leverage both high-level semantic features and morphological features in the classification process. To address this limitation, this study introduced a dual-branch network called DAFFNet, which effectively combines image morphological features with high-level semantic features, providing potential directions for future research.

The dual-branch network DAFFNet proposed in this study comprises two feature extraction modules, EPSA-SAResNet50 and MFE. The results of the MAP module in MFE for each morphological attribute are presented in Table 5 and Figure 7, demonstrating the robust performance of MAP even in case of sample imbalances. However, challenges arise in accurately predicting certain morphological attributes, such as cell size, where there is not a unified measure of cell size, and the qualitative judgement of only big and small cannot fully reflect the real size of cells. The performance of DAFFNe, as shown in Table 2 and Figure 6, reveals its superior robustness and validity across six datasets: PBC, LISC, Raabin-WBC, BCCD, LDWBC, and Labelled. Notably, DAFFNet achieves state-of-the-art results on the BCCD dataset and surpasses 98% accuracy on PBC, Raabin-WBC, LDWBC, and Labelled datasets. Despite the limited data samples of LISC (242 samples), the model's 91.30% accuracy is acceptable, underscoring its resilience and effectiveness.

In the proposed framework, ablation experiments are conducted for each module to highlight the validity of each module. For the MAP module, in order to fully utilize all the sample data and category labels, a combination of DSL and SSL training strategies is employed to maximize the utilization of sample data and category labels. . The results presented in Table 6 indicate that both DSL and SSL contribute to enhancing the performance of the MAP module, showcasing the effectiveness of integrating DSL and SSL within MFE. The decision to incorporate dual attention into ResNet50 is motivated by two key factors: emphasizing feature channels relevant for classification and capturing essential non-local information within images. From Table 4, it can be seen that the model performance after the introduction of EPSA increases from 98.62%, 97.14%, 97.86%, 97.81%, and 96.84% to 98.65%, 97.42%, 99.14%, 97.99%, and 98.18% compared to baseline on PBC, Raabin-WBC, BCCD, LDWBC, Labelled. However, in LISC it decreased from 87.23% on to 85.11%. This may be due to the fact that the images in the LISC dataset have more background information and a small proportion of leukocytes, so that EPSA cannot capture the spatial local features of the images well. On this basis, the introduction of dual attention combining EPSA with SA improves the results on all five datasets, including the LISC dataset, proving the effectiveness of the dual attention. The final dual-branch DAFFNet model effectively leverages high-level semantic information and low-level morphological features from images. The addition of the MAP and MAE modules further enhances the model's performance on all datasets, except for a minor decrease in performance on the LDWBC dataset, demonstrating the robustness and effectiveness of fusing morphological features with high-dimensional semantic features.

The integration of high-dimensional semantic features with low-level morphological features represents a significant advancement in WBC classification. While a single high-dimensional semantic feature may lack interpretability and be susceptible to model overfitting. Relying solely on morphological features for classification may not capture the intricate information in images, potentially leading to reduced accuracy. Therefore, the combination of high-dimensional semantic and morphological features holds substantial importance. Table 3 compares our results with those of recent years. Our method outperforms most of the existing methods for WBC classification on all six datasets, confirming the validity of the results and the generalization ability of the model. In addition, our model achieves 99.72% precision, 99.71% recall, 99.71% F1-score, 99.71% accuracy, and 99.82% AUC on the BCCD test set, and the overall metrics achieve state-of-the-art. which highlights the T of DAFFNet.

The method of combining morphological information with high-level semantic information for classification proposed in this study not only advances the field of leukocyte classification, but also provides new ideas for other classification tasks with morphological attribute annotations. The classification combining morphological information with high-level semantic information is highly advanced as it can fully utilize both high-level and low-level information of images. This proposed generalization shows signs of moving the classification field forward. Further research is needed to explore this approach and understand its feasibility more fully.

In the future, we plan to optimize advanced feature fusion algorithms to better combine high-level semantic features with morphological features. In addition, the use of higher quality and larger scale leukocyte images may improve the performance of the proposed technique. We also plan to build models with more generalization capabilities based on larger scale datasets after fusion of multiple datasets. To assist haematologists and physicians, we will also investigate more advanced detection and classification algorithms and develop smart and portable devices for clinical applications.

## 5. Conclusion

This study introduces a new dual-branch network called DAFFNet, which integrates morphological features and high-level semantic features of WBC for the first time in the field of WBC classification to achieve precise classification. The network enhances model accuracy and representation by incorporating dual attention mechanisms that combine EPSA and SA, as well as MFE, a feature extraction module that utilizes a combined DSL and SSL approach. Through extensive ablation and comparison experiments, the proposed model demonstrated high performance on six datasets - PBC, LISC, Raabin-WBC, BCCD, LDWBC, and

Labelled - achieving accuracy rates of 98.77%, 91.30%, 98.36%, 99.71%, 98.45%, and 98.85% respectively. These results validate the effectiveness and superiority of the proposed methodology. DAFFNet, the dual-branch network, contributes novel perspectives to the field of WBC classification and establishes a groundwork for the objective and precise detection and classification of leukocytes in microscopic blood cell images.

## Acknowledgements

The authors would like to thank all the reviewers who participated in the review. We also acknowledge the PBC, LISC, Raabin-WBC, BCCD, LDWBC, Labelled and WBCAtt datasets for providing their platforms and contributors for uploading their meaningful datasets.

## Ethical Statement

Not applicable.

## Data Availability

The datasets used in this study are all public and freely available. Codes will be made available on request.

## Funding

This research did not receive any specific grant from funding agencies in the public, commercial, or not-for-profit sectors.

## Abbreviations

**WBC** White blood cell; **CAD** Computer Aided Diagnosis; **EPSA** Efficient Pyramid Squeeze Attention; **SA** Spatial Attention; **DAFFNet** Dual Attention Feature Fusion Network; **MFE** Morphological Feature Extractor; **MAP** Morphological Attribute Predictor; **MAE** Morphological Attribute Encoder; **DSL** Deep Supervision Learning; **SSL** Semi-supervised Learning; **Prec** Precision; **Rec** Recall; **F1** F1-score; **Spec** Specificity; **Acc** Accuracy; **SAcc** Subset Accuracy; **AUC** Area Under the Curve; **HL** Hamming Loss; **JS** Jaccard Similarity